\documentclass[twocolumn,switch]{article}

\usepackage{preprint}
\usepackage[utf8]{inputenc}
\usepackage[T1]{fontenc}
\usepackage{cite}
\usepackage{amsmath,amssymb,amsfonts}
\usepackage{booktabs}
\usepackage{tabularx}
\usepackage{multirow}
\usepackage{siunitx}
\usepackage{graphicx}
\usepackage{xcolor}
\usepackage{microtype}
\usepackage{url}
\usepackage{hyperref}
\usepackage{dblfloatfix}
\usepackage{placeins}
\usepackage{tikz}
\usetikzlibrary{arrows.meta,positioning,fit}

\graphicspath{{figures/}}
\definecolor{sourceblue}{HTML}{3B6FB6}
\definecolor{targetgreen}{HTML}{2F855A}
\definecolor{filterorange}{HTML}{C05621}
\definecolor{auditred}{HTML}{B8322A}

\renewcommand{\preprintstatus}{Preprint, September 2026}

\hypersetup{
  pdftitle={Data-Efficient Crosswalk Segmentation from Overhead CCTV via Confidence- and Geometry-Guided Pseudo-Labeling},
  pdfauthor={Abdirashid Omar and Jonghyuk Park},
  pdfkeywords={crosswalk segmentation, CCTV, pseudo-labeling, semi-supervised learning, domain shift},
  colorlinks=true,
  linkcolor=black,
  urlcolor=sourceblue,
  citecolor=sourceblue
}

\newcommand{\iou}{\operatorname{IoU}}
\newcommand{\dice}{\operatorname{Dice}}
\newcommand{\repo}{\href{https://github.com/rashiedomar/crosswalk-cctv}{\texttt{rashiedomar/crosswalk-cctv}}}

\title{Data-Efficient Crosswalk Segmentation from Overhead CCTV via Confidence- and Geometry-Guided Pseudo-Labeling}

\author{
  Abdirashid Omar and Jonghyuk Park \\
  Department of Data Science, Graduate School of Kookmin University \\
  Seoul 02707, Republic of Korea
}

\begin{document}

\twocolumn[
\begin{@twocolumnfalse}
\maketitle

\begin{abstract}
Pixel-level annotation of fixed traffic-camera imagery is expensive, while crosswalk models trained from street-level imagery face a substantial viewpoint and appearance shift when applied to elevated CCTV. We investigate a data-efficient target-domain pipeline using 241 manually annotated CCTV images and 5,926 unlabeled CCTV frames. A source-domain experiment trains a 31.0M-parameter custom U-Net on 3,300 first-person-view (FPV) images and obtains 93.05\% IoU on its 330-image FPV test split. This result is a source baseline, not transferred performance: the released CCTV notebook instantiates a 42.0M-parameter DeepLabV3-ResNet50 from torchvision weights, and no compatible mapping from the U-Net checkpoint is implemented. Training on 201 manual CCTV images and selecting on 40 held-out manual masks yields 88.91\% IoU. The model then predicts all unlabeled frames; image-level certainty and a largest-component area prior rank the candidates, and the top 1,000 attain mean certainty 0.976 and mean combined score 0.988. A repository audit shows that the reported second-stage 98.52\% IoU was measured on a 150-image split containing only teacher-generated pseudo-masks: because of a directory-layout mismatch, the executed combined-data loader found zero manual samples and split 1,000 pseudo-labeled samples into 850/150. We therefore report 98.52\% as internal pseudo-label agreement rather than human-ground-truth accuracy. The defensible target-domain result is 88.91\% IoU on the 40 manual validation images. Batch-one FP32 inference at $512\!\times\!512$ requires 12.98 ms (77.03 FPS) on an NVIDIA RTX A6000 48 GB. These findings support the practicality of confidence-and-geometry filtering, while also showing why pseudo-label evaluation must remain isolated from the labels used for self-training.
\end{abstract}

\keywords{crosswalk segmentation \and CCTV \and pseudo-labeling \and semi-supervised learning \and domain shift \and intelligent transportation}
\vspace{0.25cm}
\end{@twocolumnfalse}
]

\section{Introduction}\label{sec:introduction}

Crosswalk localization is a useful perception primitive for traffic monitoring and pedestrian-safety systems. Classical methods exploit repeated stripe edges and geometric regularity \cite{ivanchenko2008crosswalk}; recent systems instead learn crosswalk appearance from data \cite{berriel2017crosswalk,liang2021zebra}. Fixed overhead cameras remain difficult because perspective compression, small foreground scale, dynamic occlusion, shadows, nighttime illumination, and camera-specific backgrounds differ sharply from pedestrian- or vehicle-level imagery.

Collecting target-domain video is easy, but dense annotation is not. This asymmetry motivates a simple question: can a small manually labeled CCTV set and a larger unlabeled target-domain pool support useful binary crosswalk segmentation under FPV-to-CCTV shift? We study that question in the public \repo{} project. The empirical pipeline trains a supervised CCTV model, predicts 5,926 unlabeled frames, rejects implausible masks with confidence and foreground-area checks, and uses the highest-ranked masks for a second training stage.

This paper is deliberately conservative. Saved notebook outputs provide strong evidence for the supervised target model and for real-time throughput, but they do not provide an independent human-annotated evaluation of the second-stage checkpoint. Our audit additionally identifies a loader mismatch that excluded all manual samples from the executed second stage. Accordingly, the paper separates three different quantities: source-domain FPV test IoU, held-out manual CCTV validation IoU, and agreement with generated pseudo-masks. Mixing them would overstate the evidence.

Our contributions are:

\begin{itemize}
  \item an empirical characterization of the FPV-to-overhead-CCTV viewpoint gap for binary crosswalk segmentation;
  \item a lightweight pseudo-label ranking rule that combines pixel certainty with a crosswalk-area prior, selecting 1,000 samples from 5,926 unlabeled frames;
  \item a reproducible audit of dataset construction and evaluation semantics, including a clear distinction between 88.91\% manual-mask IoU and 98.52\% internal pseudo-mask agreement; and
  \item a measured batch-one throughput of 77.03 FPS at $512\!\times\!512$ on an RTX A6000.
\end{itemize}

\begin{figure*}[t]
\centering
\resizebox{0.98\textwidth}{!}{%
\begin{tikzpicture}[
  node distance=4.5mm and 5.5mm,
  box/.style={draw, rounded corners=2pt, align=center, minimum height=9mm, text width=23mm, font=\footnotesize, inner sep=3pt},
  source/.style={box, draw=sourceblue, fill=sourceblue!7},
  target/.style={box, draw=targetgreen, fill=targetgreen!7},
  filter/.style={box, draw=filterorange, fill=filterorange!8},
  note/.style={draw=auditred, dashed, rounded corners=2pt, align=left, text width=52mm, font=\scriptsize, inner sep=3pt, fill=auditred!4},
  flow/.style={-{Latex[length=2mm]}, thick},
  context/.style={-{Latex[length=2mm]}, dashed, sourceblue, thick}
]
\node[source] (fpv) {3,300 FPV\\images};
\node[source, right=of fpv] (unet) {Custom U-Net\\source baseline};
\node[target, right=12mm of unet] (manual) {241 manual\\CCTV masks};
\node[target, right=of manual] (teacher) {DeepLabV3\\supervised CCTV};
\node[target, right=of teacher] (unlab) {5,926 unlabeled\\CCTV frames};
\node[filter, right=of unlab] (rank) {certainty + area\\ranking};
\node[filter, right=of rank] (pseudo) {top 1,000\\pseudo-labels};
\draw[flow, sourceblue] (fpv) -- (unet);
\draw[context] (unet) -- node[above, font=\scriptsize, align=center]{viewpoint-gap\\reference only} (manual);
\draw[flow, targetgreen] (manual) -- node[above, font=\scriptsize]{201 train} (teacher);
\draw[flow, targetgreen] (teacher) -- (unlab);
\draw[flow, filterorange] (unlab) -- (rank);
\draw[flow, filterorange] (rank) -- (pseudo);
\node[target, below=11mm of teacher] (val) {40 manual CCTV\\validation masks};
\draw[flow, targetgreen] (teacher) -- (val);
\node[target, below=11mm of pseudo] (student) {second-stage\\training};
\draw[flow, filterorange] (pseudo) -- (student);
\node[note, left=8mm of student] (audit) {Executed notebook: manual loader returned 0; the 1,000 pseudo-labeled samples were split 850/150. Thus 98.52\% measures internal pseudo-mask agreement, not independent CCTV accuracy.};
\draw[-{Latex[length=2mm]}, dashed, auditred] (audit) -- (student);
\end{tikzpicture}
}
\caption{Audited experimental pipeline. The FPV model is a source-domain reference; the released code does not implement a verified U-Net-to-DeepLab parameter transfer. Green denotes human-annotated target-domain training and validation, orange denotes pseudo-label generation, and the dashed red note records the executed second-stage protocol.}
\label{fig:pipeline}
\end{figure*}
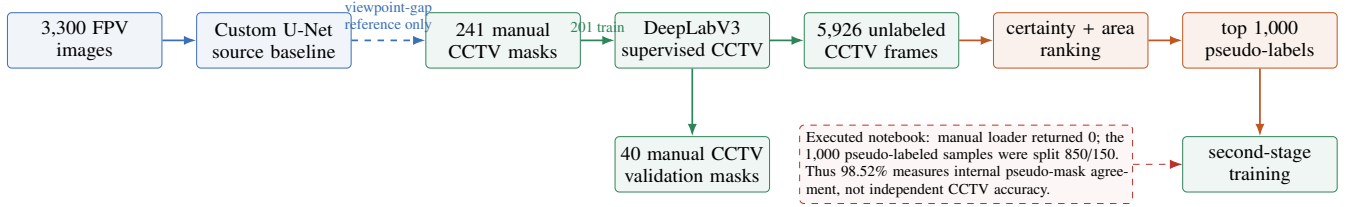

\section{Related Work}\label{sec:related}

\subsection{Crosswalk perception}

Crosswalk detection has been studied from mobile, vehicle, and overhead viewpoints. Ivanchenko et al. used line grouping and figure-ground reasoning on a camera phone \cite{ivanchenko2008crosswalk}. Berriel et al. used automatically collected street and map imagery and emphasized cross-database evaluation \cite{berriel2017crosswalk}. Liang and Seo combined SegNet-style decoding with residual features for zebra-crossing segmentation \cite{liang2021zebra}, while Verma and Ukkusuri detected crossings in satellite imagery for pedestrian-network completion \cite{verma2024satellite}. Fixed urban CCTV is neither street-level nor near-orthographic: crosswalks may be distant, oblique, partly outside the frame, or repeatedly occluded by vehicles.

\subsection{Segmentation and semi-supervision}

U-Net established an encoder-decoder design with skip connections for dense prediction \cite{ronneberger2015unet}; residual learning enabled deeper visual backbones \cite{he2016resnet}. DeepLabV3 uses atrous convolution and multi-scale context \cite{chen2017deeplabv3}, and DeepLabV3+ adds an explicit decoder \cite{chen2018deeplabv3plus}.

Pseudo-labeling treats confident predictions as targets \cite{lee2013pseudolabel}. Mean Teacher stabilizes targets through weight averaging \cite{tarvainen2017meanteacher}, while FixMatch couples thresholded predictions with strong augmentation \cite{sohn2020fixmatch}. Dense prediction has motivated segmentation-specific methods including ClassMix \cite{olsson2021classmix}, Cross Pseudo Supervision \cite{chen2021cps}, and selective self-training in ST++ \cite{yang2022stpp}. Under domain shift, AdaptSegNet aligns structured outputs \cite{tsai2018adaptsegnet}, CyCADA combines cycle-consistent adaptation \cite{hoffman2018cycada}, and DACS mixes source and target samples while using pseudo-labels \cite{tranheden2021dacs}. Our method is intentionally simpler: a single teacher ranks whole CCTV images by certainty and a binary geometry prior.

\section{Method}\label{sec:method}

\subsection{Problem formulation}

Let $\mathcal{D}_{L}=\{(x_i,y_i)\}_{i=1}^{N_L}$ denote labeled target-domain images and binary masks, and let $\mathcal{D}_{U}=\{u_j\}_{j=1}^{N_U}$ denote unlabeled CCTV frames. A network $f_\theta$ produces a foreground probability map
\begin{equation}
  p=f_\theta(x), \qquad p\in[0,1]^{H\times W},
\end{equation}
and the binary prediction is $\hat{y}_k=\mathbb{1}[p_k\geq 0.5]$.

\subsection{Source-domain baseline}

The saved FPV notebook partitions 3,300 images into 2,640 training, 330 validation, and 330 test samples. Contrary to the repository README's ``U-Net + ResNet34'' description, the executable notebook defines a custom four-level U-Net with channels 64--128--256--512, a 1,024-channel bottleneck, and 31,043,521 parameters; no ResNet encoder is instantiated. It trains for 30 epochs with Adam at $10^{-4}$, batch size 8, and the same binary objective used below. The best validation IoU is 92.44\% at epoch 21 (one-indexed), and the restored checkpoint obtains 93.05\% on the FPV test set.

The CCTV notebook separately instantiates torchvision DeepLabV3-ResNet50 with default pretrained weights and replaces the final output layer for one class. It attempts to load the custom U-Net state dictionary with \texttt{strict=False}, but does not record matched keys or implement a parameter mapping. Because the module namespaces and tensors belong to different architectures, the available implementation does not establish meaningful checkpoint transfer. We therefore treat the FPV experiment only as a source-domain baseline.

\begin{figure*}[t]
  \centering
  \includegraphics[width=0.97\textwidth]{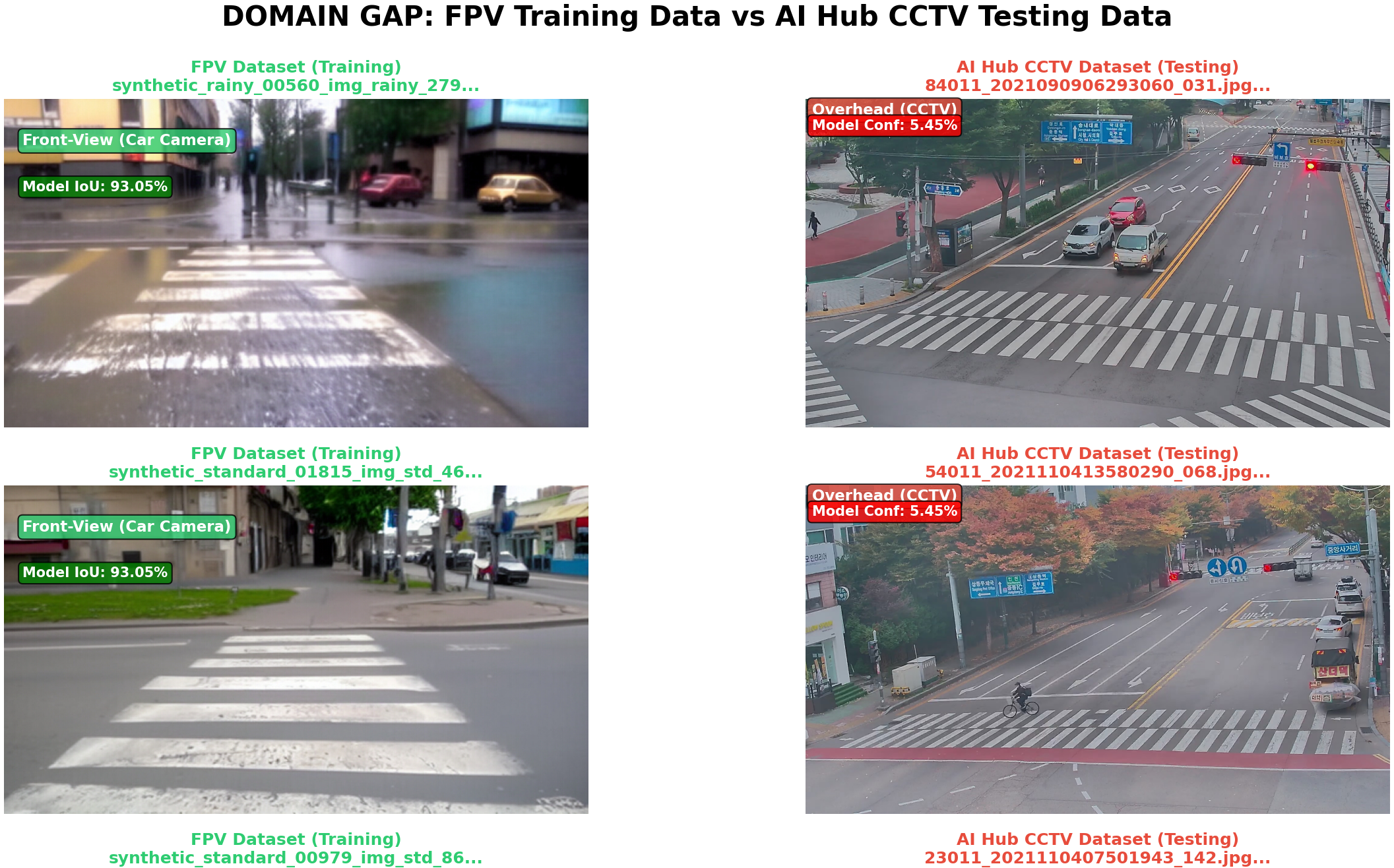}
  \caption{Examples recovered from the repository's FPV-versus-AI-Hub comparison. FPV imagery (left) is near-horizontal and crosswalk-centered; fixed CCTV imagery (right) is elevated, wider, and more cluttered. The overlaid confidence text comes from the source-model analysis and is not ground-truth IoU on CCTV.}
  \label{fig:domain}
\end{figure*}

\subsection{Supervised CCTV model}

The target model is DeepLabV3 with a ResNet50 backbone \cite{he2016resnet,chen2017deeplabv3}. Images and masks are resized to $512\!\times\!512$. The 241 human-labeled images are split with seed 42 into 201 training and 40 validation samples. The training loader uses batch size 8 and drops one incomplete sample, so 200 examples are processed per epoch. Optimization uses AdamW \cite{loshchilov2019adamw}, learning rate $10^{-4}$, weight decay $10^{-4}$, and 30 epochs.

For pixels indexed by $k$, the objective is
\begin{align}
\mathcal{L}_{\mathrm{sup}} &= \mathcal{L}_{\mathrm{BCE}}+\mathcal{L}_{\mathrm{Dice}},\\
\mathcal{L}_{\mathrm{Dice}} &= 1-\frac{2\sum_k p_k y_k+\epsilon}{\sum_k p_k+\sum_k y_k+\epsilon}.
\end{align}
The notebook applies color jitter and random horizontal flipping to training images. Its horizontal flip is not synchronized with the mask, however, and therefore adds alignment noise. This defect should be corrected in any rerun; it does not change the provenance of the archived validation score.

\subsection{Confidence- and geometry-guided selection}

For each unlabeled image, the teacher produces $p_j$ and pseudo-mask $\hat y_j$. The implementation defines image-level certainty as the mean distance from the binary decision boundary,
\begin{equation}
 c_j=\frac{2}{HW}\sum_{k=1}^{HW}|p_{j,k}-0.5|.
\end{equation}
It extracts the largest contour in $\hat y_j$, computes its area ratio
\begin{equation}
 r_j=\frac{A(\operatorname{largest}(\hat y_j))}{HW},
\end{equation}
and assigns
\begin{equation}
 g_j=\begin{cases}
 1.0,&0.05<r_j<0.40,\\
 0.5,&\text{otherwise},
 \end{cases}\qquad s_j=\frac{c_j+g_j}{2}.
\end{equation}
Candidates with $s_j\geq0.7$ are sorted, and at most 1,000 are retained. All selected samples satisfy the area rule. Their certainty is approximately 0.976, while the combined score has mean 0.9882, median 0.9882, and range 0.9879--0.9888. This distinction resolves inconsistent uses of ``confidence'' in the saved summary and README.

\begin{figure*}[t]
  \centering
  \includegraphics[width=0.94\textwidth]{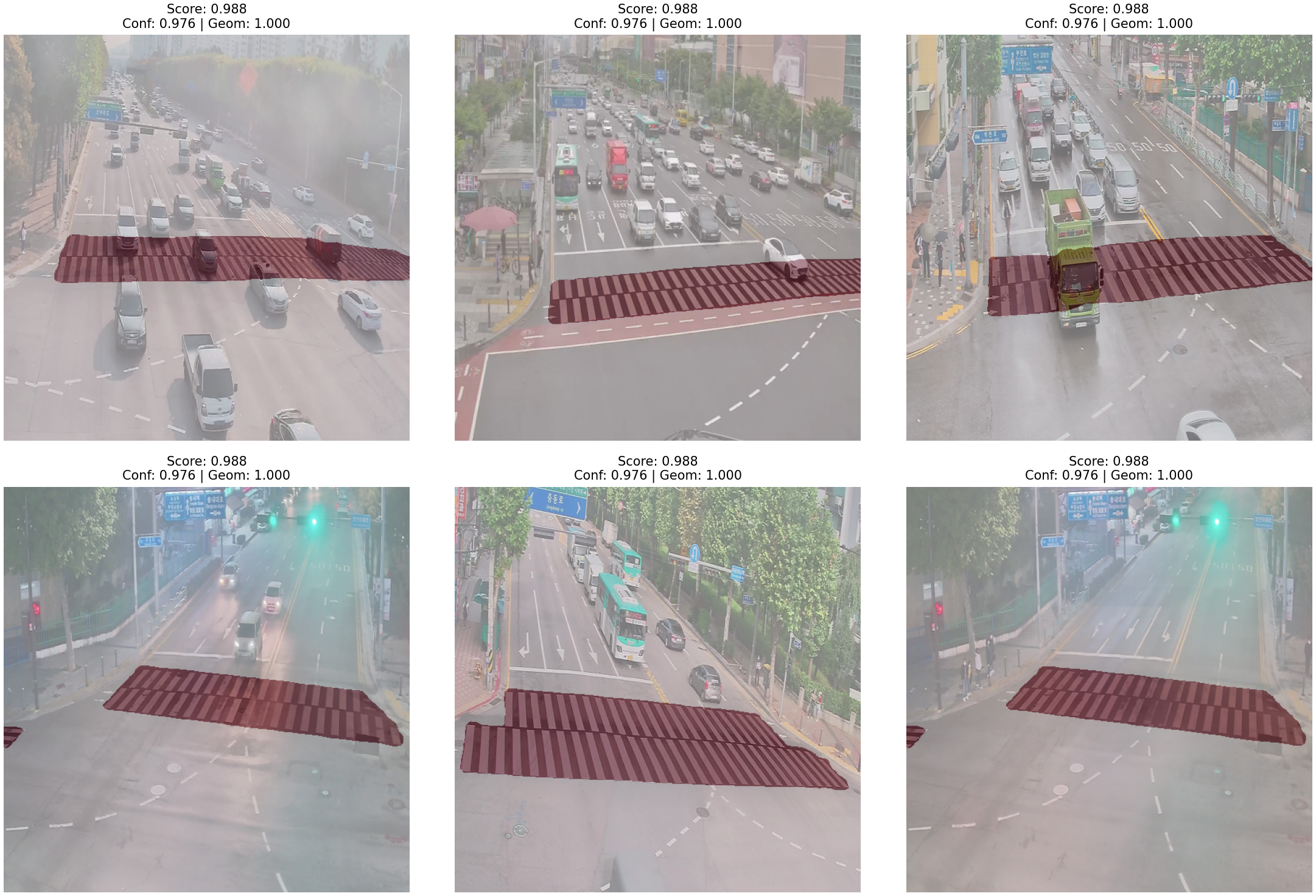}
  \caption{Six high-ranked pseudo-labels recovered from the repository. Red overlays are automatically generated masks, not human ground truth. The displayed score averages pixel certainty (about 0.976) and the binary geometry score (1.0), yielding about 0.988. Occluding vehicles remain visible beneath the translucent masks.}
  \label{fig:pseudo}
\end{figure*}

\subsection{Second-stage training and protocol audit}

The intended design combines human and pseudo-labeled data. The executed notebook does not realize that design. Its initial manual-data loader searches recursively and finds \texttt{finetuning/images} and \texttt{finetuning/masks}. The second-stage loader instead iterates over immediate subdirectories and searches for an additional \texttt{images/masks} pair below each one. It consequently prints \texttt{Original: 0}, \texttt{Pseudo: 1000}, and then randomly splits only the pseudo-labeled set into 850 training and 150 validation images. A later ``1,241 samples'' message is a hard-coded string, not the loaded dataset length.

The model continues from the supervised checkpoint and trains for 20 epochs with AdamW at $5\!\times\!10^{-5}$. Its best 98.52\% IoU at epoch 19 measures overlap with pseudo-masks generated by its own teacher lineage. It is useful as a self-training diagnostic but cannot establish an accuracy gain. The public repository excludes the images and model checkpoints, so the second-stage checkpoint cannot be reevaluated here on the 40 manual masks.

\section{Experimental Setup}\label{sec:experiments}

\subsection{Data}

The CCTV frames derive from the AI-Hub urban-road traffic CCTV resource \cite{aihub2021cctv}. Table~\ref{tab:data} distinguishes recorded data from intended and executed stage-two pools. The unlabeled frames are used only for teacher prediction and pseudo-label selection.

\begin{table}[t]
\centering
\caption{Dataset composition and executed use.}
\label{tab:data}
\small
\setlength{\tabcolsep}{3.5pt}
\begin{tabularx}{\columnwidth}{@{}Xrr@{}}
\toprule
Subset or pool & Images & Role \\
\midrule
FPV train / val / test & 2,640 / 330 / 330 & source \\
Manual CCTV train & 201 & supervised \\
Manual CCTV validation & 40 & human eval. \\
Unlabeled CCTV pool & 5,926 & teacher input \\
Selected pseudo-labels & 1,000 & stage two \\
Nominal combined pool & 1,241 & documented intent \\
Executed stage-two pool & 1,000 & 0 human + 1,000 pseudo \\
Executed train / val & 850 / 150 & pseudo only \\
\bottomrule
\end{tabularx}
\end{table}

\subsection{Metrics and implementation}

For binary masks, the reported intersection-over-union is
\begin{equation}
 \iou(\hat y,y)=\frac{\sum_k \hat y_k y_k+\epsilon}
 {\sum_k \hat y_k+\sum_k y_k-\sum_k \hat y_k y_k+\epsilon}.
\end{equation}
Dice/F1 is related by $\dice=2\iou/(1+\iou)$ \cite{dice1945}, but the archived experiments report IoU. Notebook validation averages batch-level IoUs rather than accumulating a dataset-wide confusion matrix. The manual validation set contains exactly five full batches; the pseudo-only validation set ends with a smaller batch and is therefore not strictly sample-weighted.

Experiments ran with PyTorch on an NVIDIA RTX A6000 with 48 GB VRAM. The CCTV notebook reports 41.99M trainable parameters, which we round to 42.0M; the README's ``approximately 39M'' is not consistent with the saved model printout. No mixed-precision context appears in the timing cell, so Table~\ref{tab:runtime} describes FP32 inference.

\begin{table}[t]
\centering
\caption{Recorded inference configuration.}
\label{tab:runtime}
\small
\setlength{\tabcolsep}{4pt}
\begin{tabular}{@{}ll@{}}
\toprule
Item & Value \\
\midrule
Model & DeepLabV3-ResNet50 \\
Parameters & 41.99M \\
Input / batch & $512\!\times\!512$ RGB / 1 \\
Precision & FP32 (no autocast recorded) \\
Hardware & NVIDIA RTX A6000, 48 GB \\
Warm-up / timed runs & 10 / 100 \\
Latency & 12.98 ms per image \\
Throughput & 77.03 FPS \\
\bottomrule
\end{tabular}
\end{table}

\begin{figure*}[t]
  \centering
  \includegraphics[width=0.98\textwidth]{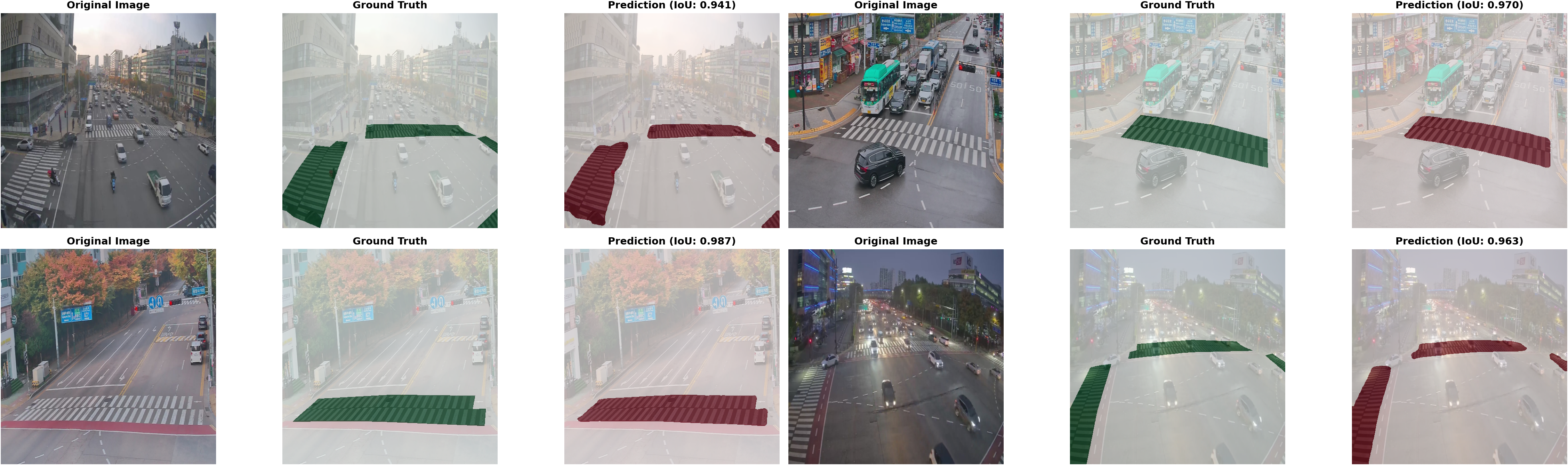}
  \par\vspace{1.5mm}
  \includegraphics[width=0.48\textwidth]{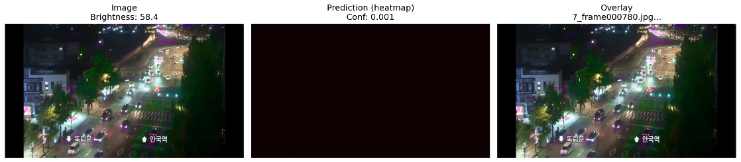}
  \caption{Qualitative evidence with provenance kept explicit. The upper grid shows four samples from the 40-image human-annotated CCTV validation set used for the supervised model: each group contains input, human ground-truth overlay (green), and prediction overlay (red). The lower strip shows the FPV source model's near-empty response to a difficult night CCTV frame; it has no human mask and is included only as a domain-shift failure example. The upper panel is not an iteration-two evaluation.}
  \label{fig:qualitative}
\end{figure*}

\section{Results}\label{sec:results}

Table~\ref{tab:results} presents each value with its actual protocol. The 93.05\% FPV test result establishes in-domain source performance. The 88.91\% CCTV result is the strongest target-domain number backed by human masks: the best of 30 epochs on the fixed 40-image validation set. Because that split also selects the checkpoint, it should be called validation, not an independent test set.

The pseudo-label statistics show that the ranking rule selects highly certain, area-valid masks. The 98.52\% second-stage value is numerically higher but answers a different question. It measures how well the continued model fits teacher-generated targets in a pseudo-only split and must not be presented as a 9.61-point target-domain improvement.

\begin{table}[t]
\centering
\caption{Results separated by label source and evaluation protocol. Values across rows are not directly comparable when the domains or targets differ.}
\label{tab:results}
\footnotesize
\setlength{\tabcolsep}{3pt}
\begin{tabularx}{\columnwidth}{@{}lXr@{}}
\toprule
Stage & Evaluation target and split & IoU \\
\midrule
FPV baseline & Human FPV masks; 330 test & 93.05\% \\
Supervised CCTV & Human CCTV masks; 40 validation & 88.91\% \\
Pseudo selection & No human targets; top 1,000 & --- \\
Stage-two internal & Teacher pseudo-masks; 150 validation & 98.52\% \\
Stage two, human CCTV & Checkpoint/data unavailable; not run & --- \\
\bottomrule
\end{tabularx}
\end{table}

Figure~\ref{fig:qualitative} confirms that the supervised CCTV model can follow several crosswalk components, including oblique boundaries, crowded scenes, and nighttime imagery. It also exposes the failure mode that motivated target-domain supervision: the FPV-only source model can return almost no foreground under distant, dark CCTV conditions. Figure~\ref{fig:pseudo} shows plausible selected masks, but visual plausibility is not a substitute for human-mask evaluation.

\FloatBarrier
\section{Discussion and Limitations}\label{sec:discussion}

The reliable conclusion is narrower than the repository README suggests. A modest manual target set is sufficient to train a strong CCTV validation model, and the teacher can generate visually plausible masks at scale. The certainty-plus-area rule is inexpensive and domain-specific: it removes masks with a largest foreground component below 5\% or above 40\% of the frame. At the same time, high certainty can reflect calibration or class imbalance rather than correctness, and an area prior cannot detect a confidently segmented road region of plausible size.

Four limitations determine the next experiment. First, 40 validation images are too few for a definitive estimate and are used for checkpoint selection. Second, the executed second stage contains no manual masks, so it is not the intended semi-supervised mixture. Third, random splitting of nearby video frames may leak scene and temporal redundancy; future splits should be camera- or sequence-disjoint. Fourth, unsynchronized image/mask flipping introduces training noise. These issues are implementation and evaluation limitations, not merely presentation details.

A corrected evaluation should (i) keep the same 40 manual frames fixed for comparison, or preferably add a camera-disjoint manual test set; (ii) train on exactly 201 manual plus 1,000 pseudo-labeled samples; (iii) apply geometric transforms jointly to image and mask; (iv) report confidence calibration and pseudo-mask quality on a manually audited subset; and (v) provide per-camera results with bootstrap confidence intervals. Ablations should compare manual-only training, all pseudo-labels, certainty-only selection, geometry-only selection, and their combination. Only such an experiment can establish whether filtered pseudo-labels improve human-ground-truth IoU.

The recorded 77.03 FPS demonstrates server-GPU feasibility, not embedded deployment. It excludes video decoding, resizing, transfer, and post-processing, and should not be extrapolated to edge devices without end-to-end measurement.

\section{Conclusion}\label{sec:conclusion}

We examined a practical pipeline for crosswalk segmentation under FPV-to-CCTV domain shift. A custom U-Net reaches 93.05\% IoU on FPV test data, while a separately initialized DeepLabV3-ResNet50 reaches 88.91\% IoU on 40 human-annotated CCTV validation images after training on 201 manual examples. Confidence and a 5--40\% largest-component area prior select 1,000 visually plausible pseudo-labels from 5,926 unlabeled frames. The reported 98.52\% second-stage score, however, is pseudo-only internal agreement because the executed loader omitted all manual samples. The study therefore supports the efficiency and real-time practicality of pseudo-label generation, but does not yet prove a human-ground-truth accuracy gain from self-training. Corrected mixed-data training and camera-disjoint manual evaluation are the essential next steps.

\section*{Data and Code Availability}

Code, notebooks, numerical summaries, pseudo-label metadata, and the experiment figures used in this paper are publicly available at \repo{}. The repository commit audited for this manuscript is \texttt{7e4a9d7f59e7de56f867a2e49ab971408e5007a0}. Large datasets and trained checkpoints are excluded by the repository's ignore rules; consequently, the final checkpoint could not be independently rerun on the manual CCTV masks from the public artifacts alone. The source CCTV resource is described by AI-Hub \cite{aihub2021cctv} and remains subject to its access and use terms.

\FloatBarrier
\bibliographystyle{ieeetr}
\bibliography{refs}

\end{document}